\documentclass[11pt,a4paper]{article}
\usepackage[utf8]{inputenc}
\usepackage{coling2020}
\usepackage{times}
\usepackage{url}
\usepackage{latexsym}
\usepackage{hyperref}
\usepackage{textcomp}
\usepackage{comment} 
\usepackage{wrapfig} 
\usepackage{tikz}
\usepackage{amsmath} 
\usepackage{booktabs} 
\usepackage{paralist} 
\usepackage{wrapfig} 
\usepackage{pdflscape}
\usepackage{colortbl}%
\usepackage{amsfonts}%
\usepackage{nopageno}
\usepackage{cancel}
\usepackage{transparent}

\usepackage{footnote}
\makesavenoteenv{tabular}

\makesavenoteenv{table}

\newcommand{\myrowcolour}{\rowcolor[gray]{0.96}}
  \usepackage[inline]{enumitem}

\usepackage{graphicx}
\colingfinalcopy 

\newcommand{\emb}{\mathit{emb}}
\newcommand{\vect}[1]{$\langle\textrm{#1}\rangle$}
\newcommand{\centertwo}[1]{\multicolumn{2}{c}{#1}}
\newcommand{\centerone}[1]{\multicolumn{1}{c}{#1}}

\title{Probing Multimodal Embeddings for Linguistic Properties: \\the Visual-Semantic Case

}

\author{%
  Adam Dahlgren Lindström,\quad Suna Bensch\footnotemark[1] ,\quad Johanna Björklund\footnotemark[1] ,\quad Frank Drewes\thanks{\quad These authors contributed equally to this work.}\\
  Department of Computing Science, Umeå University (Sweden) \\
  {\tt \{dali,\,suna,\,johanna,\,drewes\}@cs.umu.se} \\
  }

\date{}

\begin{document}
\maketitle

\begin{abstract}
        Semantic embeddings have advanced the state of the art for countless natural language processing tasks, and various extensions to multimodal domains, such as visual-semantic embeddings, have been proposed. While the power of visual-semantic embeddings comes from the distillation and enrichment of information through machine learning, their inner workings are poorly understood and there is a shortage of analysis tools. To address this problem, we generalize the notion of probing tasks to the visual-semantic case. To this end, we
        \begin{enumerate*}[label=(\roman*)]
        \item discuss the formalization of probing tasks for embeddings of image-caption pairs,
        \item define three concrete probing tasks within our general framework,
        \item train classifiers to probe for those properties, and
        \item compare various state-of-the-art embeddings under the lens of the proposed probing tasks.
        \end{enumerate*}
        Our experiments reveal an up to 12\% increase in accuracy on visual-semantic embeddings compared to the corresponding unimodal embeddings, which  suggest that the text and image dimensions represented in the former do complement each other.
        \\ 
\end{abstract}

\section{Introduction}
\blfootnote{
    %
    %

    %
    %
    %
    %
     \hspace{-0.65cm}  
     This work is licensed under a Creative Commons 
     Attribution 4.0 International License.
     License details:
     \url{http://creativecommons.org/licenses/by/4.0/}.
}

Semantic analysis aims to infer meaning from data. 
In a mathematical sense, the analysis relates objects in a syntactic domain to objects in a semantic domain. 
In natural language processing, semantic embeddings such as word2vec and, more recently, BERT and GPT-2, have had a revolutionary impact on semantic analysis. The embeddings map words to real-valued vectors which reveal semantic aspects, for example, if words are related in meaning or belong to the same topic. Creating such an embedding means to enrich as well as filter out information. Unavoidably, some (usually surface and syntactic) information will be lost in the process of projecting words and their contexts onto a representation that focuses on meaning.
Hence, there is a trade-off: the better we capture the semantics, the more surface and syntactic information becomes blurred. It depends on the downstream task what is the right balance between abstraction and detail. 

In \emph{multimodal} semantic analysis, the syntactic domain is a Cartesian product of two or more domains, such as a video with audio tracks.
This type of semantic analysis is increasingly applicable, showcasing convincing applications and results. Multimodal learning models such as DeViSE~\cite{NIPS2013_5204} demonstrate in particular that \emph{zero-shot} learning can be signficantly improved by engaging multiple modalities. There are many models based on machine learning techniques that jointly process the input modalities~\cite{ShenEtAl22019,Wu_2019_CVPR}.
The combination of modalities that has hitherto received the greatest interest is the pairing of images and text. Semantic embeddings of such data are commonly called \emph{visual-semantic embeddings}. Throughout the rest of this paper, we focus on visual-semantic embeddings, and base the empirical part of our work on the dataset \emph{Common Objects
in Context} (MS-COCO), which consists of images with captions~\cite{LinEtAl:2014}.

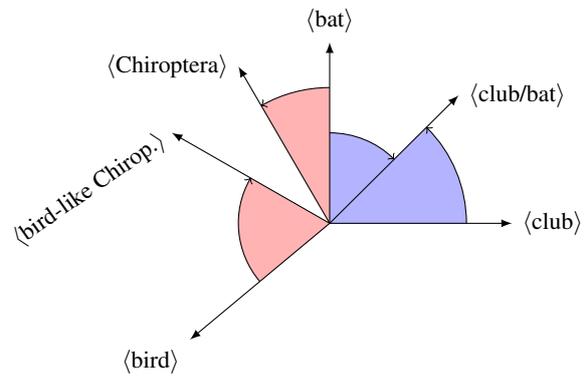
\begin{figure}
    \centering
    \begin{minipage}[t]{.23\linewidth}
    \includegraphics[trim=1 50 100 100, clip, width=\linewidth]{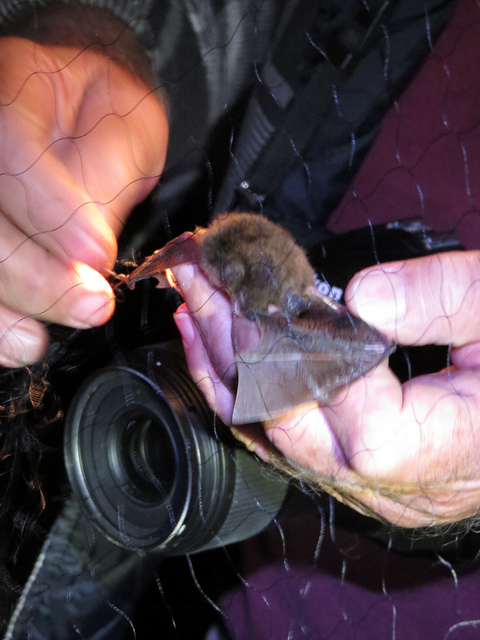}\\[1ex]
    {\small A tiny bat is held by some-\\[-.5ex] one with a camera.}
    
    {\small A man gently attempts to\\[-.5ex] feed a baby bird.}
    \end{minipage}
    \hfill
    \begin{minipage}[t]{.23\linewidth}
    \includegraphics[trim=0 13 14 16.5, clip, width=\linewidth]{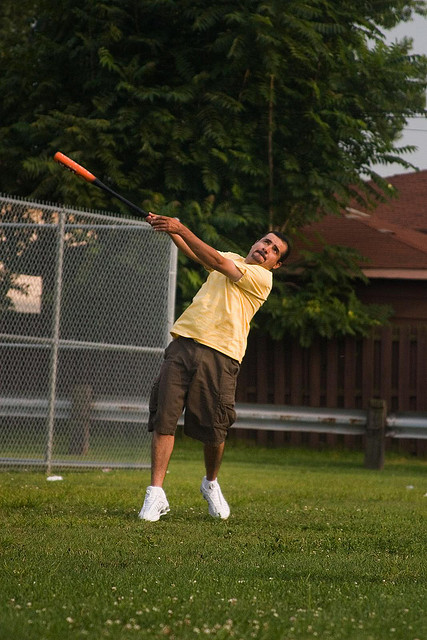}\\[1ex]
    {\small A man in shorts is swinging\\[-.5ex] a bat.}
    
    {\small A man is swinging a club\\[-.5ex] with both hands.}
    \end{minipage}
    \hfill\mbox{}
    \begin{minipage}[t]{.5\linewidth}
    \newcommand{\arc}[4]{%
      \fill[#1!30] (0,0) -- +(#2:#4) arc [start angle=#2, end angle=#3, radius=#4] -- cycle;
      \draw[->] (0,0) +(#2:#4) arc [start angle=#2, end angle=#3, radius=#4];
    }%
    \centering
    \raisebox{2cm}{\begin{tikzpicture}[scale=1.2, baseline=0cm]
      \path[use as bounding box] (-3.35cm,-1.7cm) rectangle (2.7cm,2.4cm) {};
      \arc{red}{90}{120}{1.5cm}
      \arc{red}{-140}{-210}{1cm}
      \arc{blue}{90}{45}{1cm}
      \arc{blue}{0}{45}{1.5cm}
      \draw[-latex] (0,0) -- (90:2cm) node[anchor=south] {\small \vect{bat}};
      \draw[-latex] (0,0) -- (120:2cm) node[anchor=east] {\small \vect{Chiroptera}};
      \draw[-latex] (0,0) -- (-210:2cm) node[anchor=east, rotate=35] {\small \vect{bird-like Chirop.}};
      \draw[-latex] (0,0) -- (-140:2cm) node[anchor=north east] {\small \vect{bird}};
      \draw[-latex] (0,0) -- (0:2cm) node[anchor=west] {\small \vect{club}};
      \draw[-latex] (0,0) -- (45:2cm) node[anchor=west] {\small \vect{club/bat}};
    \end{tikzpicture}%
    }
    \caption{Image-caption pairs (left) and how vectors representing the words `bat', `club', and `bird' may be affected by the image information (above)\label{fig:bat}}
    \end{minipage}
\end{figure}

When semantic analysis is applied to the text component of an image-caption pair, the visual information can resolve semantic uncertainties such as in the phrase ``a man with a bat in his hands''. Figure~\ref{fig:bat} shows two MS-COCO images,\footnote{The authors gratefully acknowledge the MS-COCO dataset~\cite{LinEtAl:2014} as the source of the two photographs, licensed under \href{https://creativecommons.org/licenses/by-nc-nd/2.0/}{CC BY-NC-ND 2.0} and \href{https://creativecommons.org/licenses/by-nc/2.0/}{CC BY-NC 2.0}, respectively. The same holds for images shown later in this article. The names of the creators and attribution parties were not supplied in the dataset. The images have been cropped to save space.} each image accompanied by two of its associated captions in the dataset. 
While humans would probably glean the correct interpretation of `bat' and `club' from the text alone (but not of `bird'), the visual-semantic information is much less ambiguous. 
The vector diagram to the right of the images illustrates how one might imagine the vectors of a word embedding such as word2vec to be affected by moving to the multimodal embedding, including visual information. Imagine the vectors \vect{bat}, \vect{bird}, and \vect{club} to be those of the pure word embedding. That is, for simplicity we assume that the words are embedded as in the original word2vec embedding, without taking the context provided by the sentence into account.
In particular, the two occurrences of `bat' in the captions are represented by one and the same vector \vect{bat}, and similarly for the two occurrences of `club'.

Now, also incorporating the information in the corresponding image may affect the vectors. The principle is shown in red for the combination of `bat' and `bird' with the left image,and in blue for the combination of `bat' and `club' with the right image. On the left, \vect{bat} becomes \vect{Chiroptera} (i.e., the vector now represents the mammal of the order Chiroptera) and \vect{bird} becomes \vect{bird-like Chiroptera}, intuitively representing a hybrid between birds and Chiropteras. In the right, both \vect{bat} and \vect{club} are turned into a vector \vect{club/bat} representing bats in the sense of clubs. Note that, while the information becomes semantically more accurate, other aspects are lost, e.g.~whether the word `bat' or `club' was used, and probably also the fact that the second caption on the left actually mentioned a bird.

Interpretable or explainable machine learning investigates ways to provide explanations for the behaviour and properties of systems based on machine learning~\cite{Gilpin2018,hase-bansal-2020-evaluating} to make them more transparent. To date, it is remains unclear what properties are encoded in semantic or visual-semantic embeddings.  
Ignorance about what is actually captured in an automatically learned semantic representation may lead to serious consequences of various kinds such as propagating discrimination bias~\cite{BolukbasiEtAl2016,CaliskanEtAl2017,BrunetEtAl2019}, or causing safety hazards in robotics by inducing unexpected robotic actions that put humans at risk \cite{Orseau2016,Wachter2017}.

This paper is intended to provide a basis for better explainability of systems relying on visual-semantic embeddings. For this, we transfer the idea of linguistic \emph{probing tasks} to this realm. The aim of probing~\cite{rogers-etal-2018-whats,ConneauEtAl:2018,yaghoobzadeh-etal-2019-probing,DBLP:journals/jair/HupkesDMB20} is to reveal what information an embedding actually encodes. Probing tasks should be agnostic to the specifics of encoder architectures, so that they can be used to compare across different methods. \newcite{ConneauEtAl:2018} define a (linguistic) probing task to be a classification task that categorizes sentences according to specific linguistic properties, such as sentence length.
From the performance of the classifier, it should be possible to draw conclusions about the probed embedding; if the classifier succeeds it indicates that the semantic embedding captures interpretable information regarding the aspect under consideration (although the converse is not necessarily true).
\newcite{HewittLiang2019} argue that the performance of a probe alone is not sufficient, and introduce so-called \emph{control tasks} to improve interpretability of probing tasks. 
A control task reveals whether high accuracy of a probing task really indicates that semantic representations encode a linguistic property, or whether the probing task itself learns this property. 
In particular, a probing task is complemented with a control task that associates random outputs to the properties under consideration (for example, POS tags).
Thus, a control task with low accuracy indicates that a corresponding probing task with high accuracy does indeed encodes the probed property.
We propose probing tasks for visual-semantic embeddings (in other words, images with captions). In particular, we are interested in tasks that shed light on whether and how a given embedding makes use of the image information in relation to linguistic phenomena such as synonyms and polysemy.

The paper is structured as follows. Section~\ref{related_work} motivates our approach and relates it to existing work. Section~\ref{sec:framework} provides a systematic discussion and formalisation of probing tasks for visual-semantic embeddings. With this, we hope to map out which properties probing tasks of various types can be used to investigate. Section~\ref{sec:concrete} introduces three concrete probing tasks that illustrate our approach, and which are used in our actual experiments reported on in Section~\ref{sec:experiments}.
The code is publicly available.\footnote{\url{https://github.com/dali-does/vse-probing}}
The conclusion in Section~\ref{sec:conclusion} summarizes our findings and lists future challenges for multimodal probing.

\section{Motivation and Related Work}\label{related_work}

The probing tasks proposed by~\newcite{ConneauEtAl:2018} probe sentence embeddings and are categorized according to the type of linguistic properties they capture: \emph{surface,- syntactic,- and semantic information}. We now give a brief account of these categories, which are outlined in Table~\ref{probing_table}. 

\begin{wraptable}{r}{.52\textwidth}
{\small
\begin{tabular}{lll}
    \toprule
    \multicolumn{3}{l}{\bfseries Type of information probed for:} \\[.5ex]
    {\bfseries Surface}  & {\bfseries Syntactic} &  {\bfseries Semantic}  \\ \midrule
    \myrowcolour%
    Sentence length & Bigram shift  &  Tense  \\
    Word content & Tree depth             &  Number of subjects  \\
    \myrowcolour%
                     & Top constituent    & Number of objects  \\
                     &             & Semantic incongruence \\
    \myrowcolour%
                      &            & Coordination inversion   \\
    \bottomrule
    \end{tabular}
}
    \caption{Probing tasks for semantic embeddings, organized along three broader probing categories as investigated in~\protect\newcite{ConneauEtAl:2018}}
    \label{probing_table}
\end{wraptable}

Surface information comprises probing for sentence length (number of words) and the word content (whether the sentence contains a given word). The probing tasks in the syntactic category ask to detect bigram shift, tree depth and top constituent, revealing whether the embedding makes certain syntactic information accessible. Bigram shift tries to predict whether two adjacent words have been swapped (that is, encoding the syntactic order of words). 
Tree depth asks to determine the depth of the syntactic tree of the sentence, and the top constituent task asks the classifier to determine the sequence of the top constituents directly below the sentence (S) node. 
The probing tasks that probe for semantic properties are tense, subject and object number, Semantic Odd Man Out, and coordination inversion. The tense task consists in finding the tense of the main verb, whereas the subject and object number tasks ask to predict the grammatical number of subjects and objects of the main verb, respectively. 
The task Semantic Odd Man Out is about predicting whether a sentence has been modified or not (i.e., a random noun or verb was replaced with another noun or verb). 
Coordination inversion probes for the information whether two coordinate clauses in a sentence have been switched. For example, ``They might be only memories, but I can feel each one'' and ``I can still feel each one, but they might be only memories'' \cite{ConneauEtAl:2018}.

These probing tasks were defined for unimodal embeddings of natural language. Machine learning that utilize multimodal embeddings is a lively field \cite{FelixEtAl2018,SocherEtAl2013,AdiKBLG17}, but little is known about what properties these multimodal embeddings actually capture. Work, such as by~\newcite{Wu_2019_CVPR}, aiming to analyze embeddings according to the composition of their encoded concepts is rare.
In this paper, we focus on probing visual-semantic embeddings.

The establishment of probing tasks is one way to gain systematic knowledge about what embeddings actually capture. Another complementary way  is to build taxonomies of multimodal machine learning techniques and multimodal embeddings. Such taxonomies are proposed, for example, by~\newcite{Baltrusaitis:19} and~\newcite{BeinbornB:18}. Both groups of authors categorize embeddings according to different but partly overlapping criteria.
The taxonomy by~\newcite{Baltrusaitis:19} classifies approaches according to five categories of criteria:
\begin{enumerate*}[label=(\alph*)]
\item \emph{representation} -- how complementary and redundant information is represented,
\item \emph{translation} -- how data is mapped between modalities,
\item \emph{alignment} -- whether and how elements in the different modalities are aligned,
\item \emph{fusion} -- how information coming from different modalities is integrated, and
\item \emph{co-learning} -- in which ways the learning exploits multimodality. 
\end{enumerate*}

The taxonomy by~\newcite{BeinbornB:18} for (learning) multimodal representations distinguishes between
\begin{enumerate*}[label=(\alph*),resume]
\item \emph{concept representations} -- embeddings that use low-level representations of concepts,
\item \emph{projections} -- embeddings that represent concepts using only one of the modalities, and
\item \emph{compositional representations} -- approaches that fuse or jointly embed the different modalities.
\end{enumerate*}

Multimodal probing tasks can support the location of a given method in a taxonomy without requiring intimate knowledge of its inner workings: probing which information is  accessible by a network trained on the resulting embeddings provides insight into what information is present and how it is represented.
Some major difficulties of multimodal processing tasks and representations are discussed (from the perspective of  multimodal grounding) by~\newcite{BeinbornB:18}. Their discussion illustrates the usefulness of multimodal probing in general, and of visual-semantic probing in particular:\footnote{We extract two aspects from the four challenges discussed by~\newcite{BeinbornB:18}, basically combining challenges~2--4, as our focus is not on grounding.}

\emph{Combining complementary pieces of information}\quad Different modalities contribute to the information content of multimodal input in complementary ways. For example, highly relevant visual properties, like the fact that birds have wings and violins are brown, are not usually mentioned in text as they are the default. Conversely, taxonomic and functional relations between concepts are poorly represented in images. Probing tasks that check whether, e.g., the word \emph{brown} relates to images of violins would allow to draw conclusions about how successfully these dimensions are combined in the embedding.

\emph{Representation of abstract concepts}\quad Multimodal grounding of verbs is difficult in comparison to grounding nouns and adjectives. This should not come as a surprise because verbs denote more abstract concepts than many nouns and adjectives do. Abstract concepts like \emph{together}, \emph{theory}, and states of mind give rise to similar difficulties. Probing tasks that evaluate how well such concepts are represented in multimodal embeddings would thus be highly useful.

\section{Systematic Probing for Properties of Visual-Semantic Embeddings\label{sec:framework}}

In this section we develop a general view of visual-semantic probing tasks, and lift the ideas of~\newcite{ConneauEtAl:2018} to the multimodal realm. Consider a property $\Pi$ that a given embedding $E$ may or may not have. In the visual-semantic case, such a property may be ``the embedding associates visual properties with the nouns in the text component'' or ``the embedding encodes the number of objects in the image''. A \emph{probing task} is defined to be a machine learning task -- usually a classification task -- that is designed in such a way that a model can be trained on $E$, and the achieved performance allows to draw conclusions regarding the extent to which $E$ possesses property $\Pi$.

We are specifically interested in developing probing tasks for visual-semantic embeddings $E$, where $\Pi$ is a property that reflects aspects of the multimodal nature of $E$. Ultimately, the goal is to come up with tasks that probe how the embedding maps the individual modalities into a common space. While we are not quite there yet, below we provide a general discussion of what to look for, and how such tasks may be categorized. Probing tasks that meet the following requirements seem to be especially valuable:
\begin{compactenum}
\item The task is a well-defined classification problem on combined (i.e., joint or coordinated) embeddings of two or more modalities.
\item The task gives insight into whether and how the multimodal embedding integrates the modalities.
\item The task has a simple and well-defined structure, so that the results are straightforward to interpret.
\item The task can be evaluated on standard data sets, or on datasets that can be created from such. 
\end{compactenum}

 We propose that the probing tasks are organized according to how they make use of the information in the sample data to map out embedding characteristics. For the visual-semantic case, at an abstract level, each probing task either probes the embedding of the original text-image pair $(T,I)$, or it is based on turning $(T,I)$ into $(T',I')$ in a well-specified manner, such that by comparing the performance of a classifier on $\emb(T',I')$  and $\emb(T,I)$, one can draw conclusions about the embedding. Depending on how $T'$ and $I'$ are obtained, different types of probing tasks arise. 

\subsection{Direct Probing}
\label{sec:identity}
Probing tasks based on $\emb(T,I)$, that is, without inflicting changes on either part, are easy to implement, but have limited potential to reveal information about the specifically multimodal characteristics of the embedding. Nevertheless, some of the probing tasks by~\newcite{ConneauEtAl:2018} have meaningful counterparts in this context. Here, we mention only the number of concepts, which is similar to sentence length and translates into \emph{complexity}: given $\emb(T,I)$, the task is to determine $|T|$, $|I|$, and $|(T,I)|$, where $|T|$ is the number of concepts mentioned in $T$ (objects and properties of objects, say), $|I|$ is the number of concepts in $I$ (i.e., the number of segments and their properties), and $|(T,I)|$ is the number of concepts in $(T,I)$. In the latter, an image segment and its counterpart in $T$ would be counted only once. Note that an embedding may be expected to be ideal for determining $|T|$ and $|I|$ if it keeps the two modalities entirely separate, while good performance on the task of determining $|(T,I)|$ indicates a tighter integration.

\subsection{Creation of Inconsistencies}
\label{sec:inconsistencies}
By considering $\emb(T-x+y,I)$ or $\emb(T,I-x'+y')$ where $y$ and $y'$ do not align with $x$ and $x'$, respectively, the effect of inconsistencies can be studied. For example, nouns in $T$ aligned to objects in $I$ may be replaced with other nouns, and similarly for adjectives referring to attributes of objects in $I$ such as position, color, size, form, and number. Variants may rely only on injecting inconsistent information, that is, $\emb(T+x,I+y)$, where $x$ and $y$ form an inconsistent pair such as $x=\mathrm{ball}$ and $y=\mathrm{cube}$. However, depending on the nature of the embedding this may require to make sure that $T+x$ is actually a reasonably well-formed sentence.

\subsection{The Challenge of Interpreting Probing Results}

We end this section with an urge for caution in the interpretation of probing task results, especially in the multimodal setting, and even more so when the results are ``negative''.

Consider the task of determining the length of the caption of a text-image pair. If classifiers trained on this task perform well, this indicates that the embedding is not well integrated. The reason is that a well-integrated embedding would blur the distinction between the image and the caption, presumably associating a high sentence length even if a complex image is provided with a short caption. Unfortunately, the converse is not true: if classifiers perform badly, the reason may equally well be that the textual part of the embedding simply does not capture sentence length, or that the chosen classifier was unsuitable for the task. 
It may thus be easier to interpret a probing task that asks for the number of \emph{objects} present in the text-image pair (see Section~\ref{sec:introspection}). 
Even in this case, poor performance does not necessarily say much about the nature of the embedding, because also a highly integrated embedding can be unsuitable for the counting task. However, despite these difficulties, this type of probing task may yield important insights if one is aware of the interpretation pitfalls. 

\section{Concrete Probing Tasks\label{sec:concrete}}
This section illustrates the abstract principles introduced in Section~\ref{sec:framework} through a set of concrete probing tasks. These tasks will be experimentally tested in Section~\ref{sec:experiments} and will, in future work, be extended with tasks of the types proposed in Section~\ref{sec:framework} to highlight complementary aspects of the semantic embeddings.
 
\subsection{Direct Probing\label{sec:introspection}}
Our first proposed probing tasks are instances of direct probing, as discussed in Section~\ref{sec:identity}: \emph{ObjectCategories} and \emph{NumObjects}. In \emph{ObjectCategories}, the task is to determine which of the 80~MS-COCO object categories are present in a given image. To turn the task into a simple classification task, we restrict the dataset to image-caption pairs in which only one of the 80~object categories is present (possibly multiple times). The second direct probing task, \emph{NumObjects}, asks to estimate the number of object instances in the image. For this task, we bin the object instances present in an image into 6 bins (5 equidistant bins for the interval 0--29, and one bin for $\geq30$ objects).

\subsection{Semantic Congruence}
\label{sec:tampering}
Detection of semantic incongruity is an example of a probing task that arises from the creation of inconsistencies (see Section~\ref{sec:inconsistencies}). It reveals whether the information propagated by $\emb(T,I)$ is sufficient to recognize that a caption has been modified, and to what extent this information stems from the visual part~$I$. The associated probing task \emph{SemanticCongruence} is the classification task that asks whether a caption has been modified. Later, we will perform this task on both $\emb(T,\emptyset)$, and $\emb(T,I)$. Without the image information, the decision must be based on purely linguistic features such as syntactic form, relative word frequencies, semantic consistency, and so forth. When the image is present, the model can also exploit incongruities between the modalities to detect modifications. 

\begin{figure}
    \centering
\begin{tabular}{@{}p{.35\linewidth}p{.65\linewidth}}
    \begin{minipage}{1.065\linewidth}
    \includegraphics[width=\textwidth]{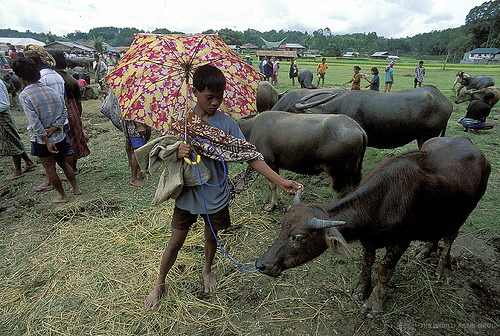} 
    \end{minipage} 
    &
    {\small
    \begin{tabular}{l}
    1.1 A \emph{child} holding a flowered umbrella and petting a yak.\\
    1.2 A \emph{checker} holding a flowered umbrella and petting a yak.\\[1ex]
    2.1 A young \emph{man} holding an umbrella next to a herd of cattle.\\
    2.2 A young \emph{mime} holding an umbrella next to a herd of cattle.\\[1ex]
    3.1 a young \emph{boy} holding an umbrella touching the horn of a cow.\\
    3.2 a young \emph{wad} holding an umbrella touching the horn of a cow.\\[1ex]
    4.1 A young \emph{boy} with an umbrella who is touching the horn of a cow.\\
    4.2 A young \emph{bear} with an umbrella who is touching the horn of a cow.\\[1ex]
    5.1 A \emph{boy} holding an umbrella while standing next to livestock.\\
    5.2 A \emph{fry} holding an umbrella while standing next to livestock.\\[1ex]
    \end{tabular}}
\end{tabular}
    \caption{In task \emph{SemanticCongruence}, the objective is to recognise semantically implausible captions.}
    \label{fig:foil}
\end{figure}

The characteristics of this probing tasks are largely determined by how the captions are modified, something that can be accomplished in numerous ways. FOIL-COCO by \newcite{Shekhar:2017} consists of modified MS-COCO pairs obtained by choosing, from each caption, a name of an object category and replacing it by another noun taken from the same MS-COCO super category. The replaced nouns occur in more than one caption, but their substitutes are salient in that they are not among the objects annotated in the image. To create plausible captions, the authors over-generate captions and use an LSTM trained on the original dataset to keep only the highest ranking ones. 

To explore a range of linguistic features broader than nouns, which are the focus of  FOIL-COCO, we compile a corpus of modified captions in which the linguistic head of each caption has been replaced. The procedure for modifying a caption works as follows. First, we run the Stanford dependency parser~\cite{QiEtAl:2020} on the caption to pick out the head. The parser also provides us with a part-of-speech tag for the head, which we use as input to the classical disambiguation algorithm by~\newcite{Lesk:1986}. The algorithm returns the most likely synonym set (synset) and the abstract category assigned to the word by Wordnet. The replacement word is picked from a synset that is in the same Wordnet category. For example, if the head is `walk' in the abstract category \emph{verb.motion} then we might choose `fly' from the same category. For simplicity, we avoid proper nouns. When the head is a verb, we prefer replacement words sharing the same set of frames, i.e., that can fill the same functions. Finally, we inflect the replacement word to match the inflection form of the head, and also mimic capitalization. 
To obtain a challenging data set, we generate $N=10$ modified sentences for each caption and then use BERT~\cite{Devlin:19} as a language model to select the best scoring alternative. This yields sentence pairs such as that of Figure~\ref{fig:foil}.

\section{Experiments\label{sec:experiments}}

This section describes our experiments with direct probing (see Section~\ref{sec:introspection}) and semantic congruence probing (see Section~\ref{sec:tampering}).
See also Appendix~\ref{sec:experiments} for further details on reproducibility.

\subsection{Experimental Setup}

\paragraph*{Dataset} We use the Microsoft Common Objects in Context (MS-COCO) dataset curated by~\newcite{LinEtAl:2014}. It consists of approximately~123\,000 images, each with at least five human-written captions. 
The object categories of the manually annotated image segments comprise 80~object categories, grouped into 11~supercategories.
We use the splits provided by~\newcite{Karpathy_2015_CVPR}, consisting of 82\,783 train, 5000 validation, and 5000 training images, respectively.
For testing, 5000 image-caption pairs over 1000 images are used of the test data, limited by what precomputed values are used by the investigated models.
This split is originally used in training all the multimodal embeddings.
We use image features precomputed by VGG19~\cite{7486599} and ResNet-152~\cite{He_2016_CVPR}, as detailed in Table~\ref{tab:capabilities}.

\paragraph*{Models}
The visual-semantic models used for our probing tasks are VSE++~\cite{VSE}, VSE-C~\cite{VSE-C}, and HAL~\cite{HAL}. In addition, we use the well-known unimodal language models BERT~\cite{Devlin:19} and GPT-2~\cite{GPT-2}.

Following the taxonomy by~\newcite{BeinbornB:18}, VSE++, VSE-C, and HAL are cross-modal transfer models trained via joint learning on the MS-COCO dataset.
The implementations of VSE-C and HAL are both based on the open source code for VSE++.
We use pretrained versions of these models, as provided with the respective papers.
VSE++ learns visual-semantic embeddings by incorporating hard negatives into the loss function and using a similarity function that scores higher for the correct image-caption pairs than for the semantically incorrect ones (that is, for the negative samples). 
VSE-C learns instead by manipulating the original captions in the MS-COCO dataset so that they constitute contrasting image-caption pairs.
HAL uses the same architecture as VSE++, but tries to avoid the so-called hubness problem where the results are skewed by frequently occurring vectors, by making the loss function aware of such structural properties of the data.

As all $X\in\{\text{VSE++}, \text{VSE-C}, \text{HAL}\}$ embed the two modalities individually (though trained on the actual multimodal data), each results in two separate models $X_\text{text}$ and $X_\text{image}$. We use these models in our experiments, in addition to ``true'' multimodal models $X_\text{avg}$ and $X_\text{conc}$ obtained by averaging and concatenating (resp.), the corresponding vectors in $X_\text{text}$ and $X_\text{image}$.

  BERT (Bidirectional Encoder Representations from Transformers) is a bidirectional language model introduced by~\newcite{Devlin:19}. It considers the left and right context surrounding a word, and relies on unsupervised learning to pre-train deep bidirectional language representations. We use an existing BERT model trained on the BookCorpus with 800 million words and on the English Wikipedia pages with 2\,500 million words~\cite{Devlin:19}. The last model, GPT-2 (Generative Pre-Training, second generation), is a transformer-based unidirectional language model trained on 40GB of lightly curated Internet text~\cite{GPT-2}.
  We use the Transformers library for these models~\cite{Wolf2019HuggingFacesTS}.
  
 \begin{wraptable}[10]{r}{.55\textwidth}
\vspace*{-1em} 
\small
\begin{tabular}{lrrr}
    \toprule
         \textbf{Model}& \textbf{Precomputed features}& \textbf{Emb. Size} & \textbf{Parameters}    \\
         \midrule
        \myrowcolour%
        VSE++ & VGG19 & 1024 &15.5(159.2)M \\
        VSE-C &  ResNet-152& 1024 & 13.8(74.1)M \\
        \myrowcolour%
        HAL  &  ResNet-152 & 1024 &11.3(71.6)M \\
        GPT-2  & &768 & 117M \\
       \myrowcolour%
        BERT  & &768 & 110M \\
        \bottomrule
    \end{tabular}
\caption{Overview of the investigated embeddings. The total size of the model, including models used to extract precomputed image features, is given in parenthesis.\label{tab:capabilities}}
\end{wraptable}

\paragraph*{Probing} 
We perform the three classification tasks \emph{ObjectCategories}, \emph{NumObjects}, and \emph{SemanticCongruence}. For \emph{NumObjects}, the label distribution between the 6 bins (see Section~\ref{sec:introspection}) is 47\,443, 17\,580, 9\,626, 4\,549, 2\,061, 1\,524 during training and 3\,025, 1\,060, 470, 240, 130, 75 during testing.
Our baseline is naively guessing the largest class. The \emph{ObjectCategories} task is based on 9\,629 and 1\,145 samples in the training and test data, respectively.
For the \emph{SemanticCongruence} task, a modified caption is chosen with probability $0.5$, and remains the same for all models tested for fair comparison.
We note that more complex models could yield higher accuracies, but following the results of~\cite{HewittLiang2019} on probe model selectivity, this improvement does not necessarily reflect the availability of the information probed for. 
Therefore, we use two classifiers for probing; one multilayer perceptron (MLP) with one hidden layer of 256 nodes and sigmoid activation, and one linear classifier with softmax activation.  Both models use a dropout of 0.2, similar to~\cite{ConneauEtAl:2018}.  
The probing models are trained on the MS-COCO data for 30 epochs using the cross-entropy loss function.
In all cases the models start to converge within the last 10 epochs.
The results reported are for the test split.

\paragraph*{Embeddings}

For each probing task, the input to the classifier is either the image embedding from one of our used models ($\text{VSE++}_\text{image}$, $\text{VSE-C}_\text{image}$, $\text{HAL}_\text{image}$) or the text embedding from one of our models ($\text{VSE++}_\text{text}$, $\text{VSE-C}_\text{text}$, $\text{HAL}_\text{text}$, BERT, GPT-2). 
For the size of the embeddings we refer to Table~\ref{tab:capabilities}.
For \emph{SemanticCongruence} the $X_\text{text}$ input consists also of modified captions. 
In order to contrast the probing results obtained with those for embeddings containing the full visual-semantic information, we also consider $X_\text{avg}$ and $X_\text{conc}$, for $X\in\{\text{VSE++}, \text{VSE-C}, \text{HAL}\}$.
All weights of each model are frozen, meaning that no weights are updated for the embedding models during the probing.

\subsection{Results and Analysis}
\begin{wraptable}[26]{r}{.545\textwidth}
\vspace*{-2.9em}
\centering
\small
\tabcolsep=1ex
    \begin{tabular}{lllcllcll}
    \toprule
    \textbf{Embedding} & 
    \centertwo{\textbf{ObjectCat.}} & &
    \centertwo{\textbf{NumObjects}} & &
    \centertwo{\textbf{SemanticCon.}}\\
    & \centerone{MLP} & \centerone{lin} && \centerone{MLP} & \centerone{lin} && \centerone{MLP} & \centerone{lin}\\
    \midrule
    \myrowcolour
    \emph{Baseline} & \centertwo{-} && \centertwo{0.605} && \centertwo{0.502}\\
    \midrule
    \emph{Image} &  &  & &  &  & & & \\
    \myrowcolour
    $\text{VSE++}_\text{image}$ & 0.753 & 0.768 && 0.646 & 0.613 && 0.502 & 0.506\\
    
    $\text{VSE-C}_\text{image}$ & 0.754 & 0.675 && 0.654 & 0.629 && 0.503 & 0.504\\
    
    \myrowcolour
    $\text{HAL}_\text{image}$ & 0.799 & 0.730 && 0.674 & 0.633 && 0.533 & 0.510 \\
    
    \midrule
    \emph{Text} &  &  && & && & \\
    
    \myrowcolour
    $\text{VSE++}_\text{text}$ & 0.862 & 0.863 && 0.627 & 0.610 && 0.739 & 0.710\\
    
    $\text{VSE-C}_\text{text}$ & 0.838 & 0.805 && 0.629 & 0.617 && 0.763 & 0.756\\
    
    \myrowcolour
    $\text{HAL}_\text{text}$ & 0.826 & 0.648 && 0.625 & 0.611 && 0.730 & 0.737\\
    
    BERT & 0.878 & 0.365 && 0.622 & 0.599 && \emph{0.816} & \emph{0.768\rlap{\footnotemark}}\\
    
    \myrowcolour
    GPT-2 & 0.811 & 0.137 && 0.617 & 0.585 && \textbf{0.792} & 0.718 \\
    
    \midrule
    \emph{Merged} &  &  && & && & \\
    
    \myrowcolour
    $\text{VSE++}_\text{avg}$ & 0.862 & 0.876 && 0.658 & 0.638 && 0.707 & 0.662\\
    
    $\text{VSE++}_\text{conc}$ & \textbf{0.911} & \textbf{0.901} && 0.661 & 0.641 && 0.743 & 0.713\\
    
    \myrowcolour
    $\text{VSE-C}_\text{avg}$ & 0.831 & 0.783 && 0.665 & 0.636 && 0.735 & 0.713\\
    
    $\text{VSE-C}_\text{conc}$ & 0.896 & 0.879 && 0.666 & \textbf{0.652} && 0.776 & \textbf{0.758}\\
 
    \myrowcolour
    $\text{HAL}_\text{avg}$ & 0.847 & 0.820 && 0.667 & 0.642 && 0.712 & 0.702\\
    
    $\text{HAL}_\text{conc}$ & 0.903 & 0.849 && \textbf{0.683} & 0.648 && 0.730 & 0.730\\
    \midrule
    \emph{Improvement} & & && & &&  & \\
    \emph{by merging} & & && & &&  & \\
    
    $\text{VSE++}$ & 0.049 & 0.038 & & 0.015 & 0.028 & & 0.040 & 0.003\\
    
    \myrowcolour
    $\text{VSE-C}$ & 0.058 & 0.074 & & 0.012 & 0.023 & & 0.013 & 0.002\\
    
    $\text{HAL}$ & 0.077 & 0.119 & & 0.009 & 0.015 & & 0.000 & \llap-0.007\\
    \bottomrule
    \end{tabular}
    \caption{Probing accuracies using a MLP with embeddings as input. The bottom three show for each model the difference between the best unimodal and the best merged embedding. All results are averaged over 5 runs and have variance $\leq0.01$.\label{tab:results}}
\end{wraptable}

Table~\ref{tab:results} shows the results for the \emph{ObjectCategories}, \emph{NumObjects} and \emph{SemanticCongruence} tasks, using both a MLP and a linear probe. While the two types of probes perform differently, the relative behavior across embeddings is very similar. A notable deviation from this general rule is the performance of BERT and GPT-2 on the \emph{ObjectCategories} task (see below).

\footnotetext{Since BERT is used during the generation of congruencies, this result is somewhat self-referential.}

\paragraph*{ObjectCategories}

We note that the text-only embedding for all three visual-semantic models yields better performance on the \emph{ObjectCategories} task than the corresponding text-only embedding, with the exception of the linear probe for HAL.
Worth noting is that out of the multimodal embeddings, HAL performs well on image-only but worst on text-only for both probes.
Further, there is a large gap between the performances of MLP and linear probes on the BERT and GPT-2 embeddings. 
This supports the conclusion of \newcite{HewittLiang2019} that MLPs, rather than acting as probes, may simply learn the task itself if provided with sufficiently rich embeddings as input, and that, therefore, linear probes may be a more appropriate choice.

Note also that BERT performs best for both probes in the text-only case, while GPT-2 scores the lowest.
All merged embeddings significantly outperform their corresponding unimodal embeddings, with concatenated VSE++ scoring the highest for both probes.
Merging the embeddings shows an improved accuracy of 3.8--11.9\% across both probe types, which suggests that the visual-semantic models combines the multimodal data in a useful way to capture which objects are present in a scene.
Overall \emph{VSE++} seems to best capture and combine information about the object categories, beating BERT and GPT-2 by a large margin for both probes.

\paragraph*{NumObjects}
The results for the \emph{NumObjects} task show that the text embeddings consistently encode the probed information in a less accessible manner than the corresponding image embeddings which are, in turn, outperformed by their merged counterparts. Using MLP probing, HAL reaches the highest accuracy on both image-only and its merged embeddings, whereas VSE-C appears to be on par with HAL on merged embeddings under a linear probe, the precise result depending on the merging strategy.
It is worth noting that the improvements from merging the embeddings are small, but are larger when using a linear probe. Once again, this supports the conclusion of \newcite{HewittLiang2019} as it indicates that the weaker probes exhibit a better sensitivity.

It is worth noting that the best result for the \emph{NumObjects} task is only about 8\% better than the baseline. This seems to indicate that the task could be improved. Most of the images contain
fewer than 10~object instances, thus falling into classes~1 and~2.\footnote{Remember that we have 6 output labels representing the number of object instances.}
The per-class accuracy numbers (Table~\ref{tab:per label} in the appendix) show that the accuracy for most embeddings and models is above 90\% for class~1, and between 30-50\% for classes~2,3, and 6.
Classes~4 and~5 (i.e., 18--23 and 24--29 object instances) yield accuracies of approximately 4--18\% and 3--15\%, respectively.
Further, the per-class accuracies show that the linear probes show performance comparable to the MLP probe on the first three classes, but never learn the 24--29 object class, and very few  of the 18--23 and $\geq30$ samples.

Image scenes containing 0--5 object instances can exhaustively be described with words, mentioning numbers and listing distinct objects explicitly (``a cup and a fork''), whereas scenes containing 18--29 objects are harder to explicitly describe. The high accuracy for scenes with more than 29 objects may be due to the fact that the large number of object instances is a ``property of the image'' and might therefore be described with words such as ``crowd''. A more balanced distribution could amplify the differences.

\paragraph*{SemanticCongruence} 
The results obtained from the \emph{SemanticCongruence} probing suggest that the additional information provided by the multimodal component does not make up for the relative loss of linguistic information.
This becomes particularly clear when using linear probing.
VSE-C$_\text{text}$ outperforms VSE++$_\text{text}$ and HAL$_\text{text}$, but is in turn clearly outpaced by the unimodal embeddings BERT and GPT-2.
If we add visual information (to VSE++, VSE-C, and HAL), the performance generally does not increase, and even decreases in one instance.
Our interpretation is that the alternative captions can be recognized from linguistic patterns such as verb-preposition agreement and other contextual information soley from having a good language understanding.
Going back to Figure~\ref{fig:foil}, we recognize that a well-formed sentence can still be highly unlikely given an understanding of language, just as Chomsky's famous example ``Colorless green ideas sleep furiously''~\cite{Chomsky:75}. Further, although the visual information could provide additional clues, it also adds noise and makes the relative proportion of linguistic data smaller. 
If this interpretation is correct, an improved linguistic quality of the alternative sentences should make the visual information more valuable for the task. 
Since this visual information cannot encode whether the caption was modified, HAL$_{\text{image}}$ aligns with the results of~\newcite{HewittLiang2019}, suggesting that this MLP probe learns something other than the probing task.
Finally, we note the good performance of BERT despite the fact that BERT was the embedding used to select the most convincing alternative captions, which should make them particularly apt at confusing BERT.

\paragraph*{Summary} 
We see that the multimodal embeddings in the \emph{merged} section of Table~\ref{tab:results} outperform their image- and text-only embeddings on the tasks \emph{ObjectCategories} and \emph{NumObjects}.
This indicates that the text- and image-only embeddings complement each other in what information they encode, and that merging them can utilize this fact.
The concatenated embeddings yield consistently better performance than the averaged ones, probably because the complementary information is fully retained.
It is not clear how well the text- and image-only embeddings project to the same space, which together with the introduction of noise from the respective modality can cause averaging to drown out important information.
Still, averaging gives better performance than unimodal approaches except for \emph{VSE-C} on \emph{ObjectCategories}.
The first two tasks are highly visual, which makes it only reasonable that the image embeddings encode more information of concern in these problems.
It is also suggested from the results that the state-of-the-art unimodal text embeddings have a better semantic language understanding.
It seems that there is a trade-off between language modeling versus understanding visual concepts, and that the training of the multimodal models has favored the latter.
This idea also aligns with the fact that these models are built for image-to-text and text-to-image retrieval, a task for which the unimodal embeddings are insufficient.
Interestingly, HAL seems to be more focused on visual information as seen in the results on \textit{ObjectCategories} and \textit{NumObjects}.
This could help explain why HAL outperforms VSE++ and VSE-C on text-to-image and image-to-text retrieval.
We also note, importantly, that the language models are larger by factors 1.45 up to 10, excluding/including the network used to precompute image features, respectively.
This can help explain why the multimodal models are not as capable in distilling the probed information in the text-only \textit{ObjectCategory} task as BERT.
To conclude, the results show that the image and text embeddings complement each other in understanding visual concepts, but that this does not extend to the understanding of language itself, as shown in the results on \emph{SemanticCongruence}.
Therefore, we conjecture that there is significant room for improvement on the multimodal embeddings for understanding scene semantics.

\section{Conclusions and Future Work}
\label{sec:conclusion}
Probing semantic embeddings with neural-network based classifiers is like looking into a black box with a lens that is itself a black box. Valuable information can still be derived, but experiments that take this approach must be made with care, and the results analysed with caution.  One approach to mitigate such opacity is proposed by~\newcite{HewittLiang2019}, namely that the probing task is complemented with a control task to alleviate a possible misinterpretation of what semantic representations actually encode.

In the multimodal setting, it is helpful to use probing tasks (as well as complementing control tasks) that are simple, well-defined, and easily implemented on standard data sets. The importance of a task being well-defined is illustrated, albeit in a negative way, by the \emph{NumObjects} task: Since there are countless equally valid ways to semantically decompose an image, it can simultaneously be true that an image shows dozens of sheep and that it shows a single herd. The flaw is arguably not as much in the task itself, as in the combination of task and data set. We may, for example, expect that the \emph{NumObjects} task comes to its right in situations where logical units of counting are understood in advance, e.g., in the case of camera footage tracking traffic congestion, where a natural unit would be the number of vehicles. 

An interesting finding from our initial experiments was the importance of linguistic compared to visual information for complexity estimation and semantic incongruity detection. 
As our next steps, we would like to repeat the experiments presented here on multimodal data sets other than the original MS-COCO dataset, in particular the Flickr~\cite{flickr30k} and FOIL-COCO~\cite{Shekhar:2017} datasets. The semantic congruity task can be implemented in numerous different ways, which makes it a particularly versatile tool for understanding language grounding. Continued work will explore how visual-semantic embeddings respond to different types of linguistic manipulation of both image captions, as well as direct image manipulation. 

\subsection*{Acknowledgements}
This work has been funded by Eureka and the Swedish Governmental Agency for Innovation Systems through the Eurostars program, grant agreement no.~11776.
We thank the anonymous reviewers for their valuable feedback, which has had a substantial influence on the final version of the paper.

\bibliographystyle{coling}
\bibliography{COLING2020}

\clearpage

\pagestyle{empty}

\appendix

\section{Experiment Details}
\label{app:exp}

The implementation is written in Pytorch 1.4.0 and trained on a NVIDIA Tesla V100 32GB GPU using CUDA 10 with Tensorflow 2.1.
The models are all trained for 30 epochs, where each epoch times in at 100 seconds on average, and the experiments are conducted using the Adam optimizer with learning rate $1.0 \times 10^{-4}$ for \emph{ObjectCategories} and \emph{NumObjects}, and $1.0 \times 10^{-3}$ for the \emph{SemanticCon} probing task.
During initial experiments SGD was also considered but Adam showed better performance.

The implementations of VSE++\footnote{\url{https://github.com/fartashf/vsepp}}, VSE-C\footnote{\url{https://github.com/vacancy/VSE-C}}, and HAL\footnote{\url{https://github.com/hardyqr/HAL}} the open sourced Github repositories with the best corresponding pretrained models are used.
For BERT and GPT-2, the Python library Transformers\footnote{\url{https://github.com/huggingface/transformers/}} is used to access pretrained models.
In both cases the base model is used, since initial experiments showed no significant difference when using larger models and in interest of keeping the comparison fair given that the larger models are substantially larger than the visual semantic embedding models.
All pretrained models are outlined in Table~\ref{tab:pretrained:models}.
Random numbers generated with \texttt{Numpy} uses a fixed seed of 1974, to make the experiments reproducible.

\begin{table}[h]
\centering
\small
    \begin{tabular}{ll}
    \toprule
    \textbf{Model} & \textbf{Version} \\
    \midrule
    \myrowcolour
    $\text{VSE++}_\text{image}$ & \texttt{runs/vsepp/runs/coco\_vse++} -\\
    
    $\text{VSE-C}_\text{image}$ & \texttt{runs/coco\_noun/ model\_best.pth.tar}\\
    
    \myrowcolour
    $\text{HAL}_\text{image}$   &  \texttt{runs/COCO\_resnet\_best\_model\_463.2.pth.tar}\\
        
    BERT & \texttt{bert-base-uncased}\\
    
    \myrowcolour
    GPT-2   &  \texttt{gpt2}\\
    \bottomrule
    \end{tabular}
\caption{The pretrained models for each of the investigate models\label{tab:pretrained:models}}
\end{table}

\section{Probing Task Per-Label Accuracies}

Table~\ref{tab:per label} gives a more detailed account of the accuracy of the tested models for the task \emph{NumObjects}. The class labels correspond to the number of objects annotated in the image.

\begin{table}[h]
\centering\small
\tabcolsep=1ex
    \begin{tabular}{lcllcllcllcllcllcll}
    \toprule
    \textbf{Model} && \centertwo{\textbf{0--5}}&& \centertwo{\textbf{6--11}}&& \centertwo{\textbf{12--17}}&& \centertwo{\textbf{18--23}}&&\centertwo{\textbf{24--29}}&& \centertwo{\textbf{\mathversion{bold}$\geq$30}}\\
    && \centerone{MLP} & \centerone{lin} && \centerone{MLP} & \centerone{lin} && \centerone{MLP} & \centerone{lin} && \centerone{MLP} & \centerone{lin} && \centerone{MLP} & \centerone{lin} && \centerone{MLP} & \centerone{lin}\\
    \midrule
    \emph{Image}\\
    \myrowcolour
    $\text{VSE++}_\text{image}$ && 0.918&0.946 && 0.322&0.227&& 0.252&0.085&& 0.192&0.004&& 0.120&0.000&& 0.400&0.184\\
    
    $\text{VSE-C}_\text{image}$ && 0.923&0.928&& 0.397&0.338 &&0.280&0.254 && 0.062&0.073&& 0.000&0.000 && 0.480&0.533\\
    
    \myrowcolour
    $\text{HAL}_\text{image}$  &&0.909&0.966&& 0.414&0.209&& 0.330&0.021&& 0.154&0.000&& 0.040&0.000&& 0.533&0.000 \\
    
    \midrule
    \emph{Text}\\
    
    \myrowcolour
    $\text{VSE++}_\text{text}$  && 0.927&0.926&& 0.287&0.256&& 0.153&0.090&& 0.039&0.038&& 0.000&0.000 && 0.240&0.133\\
    
    $\text{VSE-C}_\text{text}$ && 0.920&0.962 &&  0.390&0.218&&  0.287&0.049 &&0.105&0.000 &&0.002&0.000 &&0.560&0.187 \\
    
    \myrowcolour
    $\text{HAL}_\text{text}$  &&0.937&0.958&& 0.264&0.148&& 0.124&0.000&& 0.085&0.000&& 0.024&0.000&& 0.227&0.000  \\
    
    BERT &&0.961&0.992&& 0.162&0.040&& 0.132&0.000&& 0.000&0.000&& 0.032&0.000&& 0.200&0.000\\
    
    \myrowcolour
    GPT-2 &&0.921&1.000&& 0.216&0.000&& 0.230&0.000&& 0.038&0.000&& 0.054&0.000&& 0.240&0.000\\
    
    \midrule
    \emph{Merged}\\
    
    \myrowcolour
    $\text{VSE++}_\text{avg}$  && 0.909&0.944 && 0.380&0.311 && 0.301&0.136&& 0.115&0.038&& 0.056&0.000&& 0.427&0.240\\
    
    $\text{VSE++}_\text{conc}$  && 0.930&0.940&& 0.375&0.315&& 0.310&0.165 && 0.181&0.054&& 0.152&0.000&& 0.320&0.240\\
    
    \myrowcolour
    $\text{VSE-C}_\text{avg}$  && 0.928&0.963 && 0.337&0.281 && 0.254&0.101 && 0.077&0.020 && 0.000&0.000 && 0.533&0.253\\
    
    $\text{VSE-C}_\text{conc}$  && 0.927&0.951  && 0.294&0.336 && 0.166&0.139 && 0.040&0.069&&  0.008&0.000 && 0.251&0.280\\
          
    \myrowcolour
    $\text{HAL}_\text{avg}$  &&0.925&0.967&& 0.402&0.265&& 0.254&0.163&& 0.127&0.015&& 0.080&0.000&& 0.507&0.077\\
    
    $\text{HAL}_\text{conc}$ &&0.920&0.947&& 0.430&0.329&& 0.303&0.223&& 0.173&0.000&& 0.112&0.000&& 0.520&0.000 \\
    \bottomrule
    \end{tabular}
\caption{Accuracy per label of the tested models for the task \emph{NumObjects}\label{tab:per label}.}
\end{table}

\clearpage
\section{Semantic Congruence Dataset}
This section gives a sample of images from MS-COCO, together with original captions $(x.y.1)$, where $x \in \mathbb{N}$ identifies the image and $y \in \{1, \ldots, 5\}$ identifies the original caption, and two series of modified versions, one series $(x.y.2)$ of lower quality, and one series $(x.y.3)$ of higher quality, both modifying the caption $x.y$. We use the higher quality series in the semantic congruence task. The modified versions illustrate some of the challenges of automatically generating syntactically valid alternatives. The most common reason for poor captions is that lexical disambiguation has failed, or that we do not control for verb-preposition coherence.

    \begin{minipage}[t]{.4\textwidth}\vspace{0pt}%
    \includegraphics[height=\textwidth]{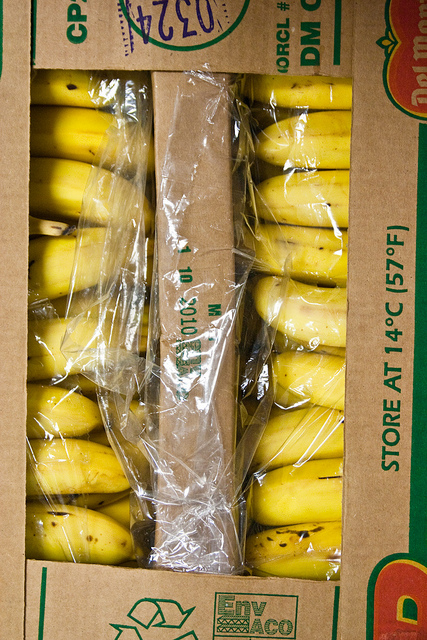} 
    \vfill
    \end{minipage} 
    \begin{minipage}[t]{.58\textwidth}\vspace{0pt}%
    {\small
    \begin{tabular}{l}
    1.1.1 This is a \emph{case} full of yellow bananas.\\
    1.1.2 This is a \emph{giant} full of yellow bananas.\\
    1.1.3 This is a \emph{squirt} full of yellow bananas.\\[1ex]
    1.2.1 \emph{Bananas} a tightly packed and boxed for delivery to the market.\\
    1.2.2 \emph{Ivy} a tightly packed and boxed for delivery to the market.\\
    1.2.3 \emph{Bunt} a tightly packed and boxed for delivery to the market.\\[1ex]
    1.3.1 a bunch of bananas are \emph{packed} up in boxes.\\
    1.3.2 a bunch of bananas are \emph{laced} up in boxes.\\
    1.3.3 a bunch of bananas are \emph{grassed} up in boxes.\\[1ex]
    1.4.1 An unopened \emph{box} full of perfectly ripe bananas.\\
    1.4.2 An unopened \emph{imperial} full of perfectly ripe bananas.\\
    1.4.3 An unopened \emph{shred} full of perfectly ripe bananas.\\[1ex]
    1.5.1 \emph{Bananas} packed in cardboard box covered in plastic.\\
    1.5.2 \emph{Paste} packed in cardboard box covered in plastic.\\
    1.5.3 \emph{Wad} packed in cardboard box covered in plastic.\\

    \end{tabular}
    }
    \end{minipage}
    
    \vspace{4ex}

    \begin{minipage}[t]{.4\textwidth}\vspace{0pt}%
    \includegraphics[width=\textwidth]{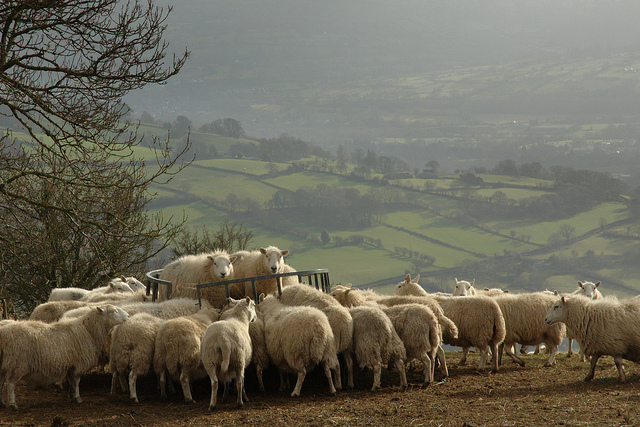} 
    \end{minipage} 
    \begin{minipage}[t]{.58\textwidth}\vspace{0pt}%
    {\small
    \begin{tabular}{l}
    2.1.1 That \emph{looks} like a wall mural in the background of this photo ...\\
    2.1.2 That \emph{occupies} like a wall mural in the background of this photo ...\\
    2.1.3 That \emph{runs} like a wall mural in the background of this photo ...\\[1ex]    
    2.2.1 A \emph{flock} of sheep who are standing a top a mountain \\
    2.2.2 A \emph{session} of sheep who are standing a top a mountain.\\
    2.2.3 A \emph{board} of sheep who are standing a top a mountain.\\[1ex]
    2.3.1 A huge heard of sheep are all \emph{scattered} together.\\
    2.3.2 A huge heard of sheep are all \emph{pumped} together.\\
    2.3.3 A huge heard of sheep are all \emph{resurfaced} together.\\[1ex]
    2.4.1 There is a \emph{gathering} of sheep in the field.\\
    2.4.2 There is a \emph{gathering} of sheep in the field.\\
    2.4.3 There \emph{litters} a gathering of sheep in the field.\\[1ex]
    2.5.1 A \emph{group} of white sheep in grassy area next to trees.\\
    2.5.2 A \emph{brass} of white sheep in grassy area next to trees.\\
    2.5.3 A \emph{mash} of white sheep in grassy area next to trees.\\
    \end{tabular}
    }
    \end{minipage}
    
    \vspace{4ex}

    \begin{minipage}[t]{.4\textwidth}\vspace{0pt}%
    \includegraphics[width=\textwidth]{Images/000000219578.jpg} 
    \end{minipage} 
    \begin{minipage}[t]{.58\textwidth}\vspace{0pt}%
    {\small
    \begin{tabular}{l}
    3.1.1 A \emph{dog} and cat lying together on an orange couch.\\
    3.1.2 A \emph{portion} and cat lying together on an orange couch.\\
    3.1.3 A \emph{wad} and cat lying together on an orange couch.\\[1ex]
    3.2.1 A dog and a cat \emph{curled} up together on a couch.\\
    3.2.2 A dog and a cat \emph{banged} up together on a couch.\\
    3.2.3 A dog and a cat \emph{hurtled} up together on a couch.\\[1ex]
    3.3.1 A \emph{cat} and dog napping together on the couch.\\
    3.3.2 A \emph{tooth} and dog napping together on the couch.\\
    3.3.3 A \emph{brush} and dog napping together on the couch.\\[1ex]
    3.4.1 A dog and cat are \emph{sleeping} together on an orange couch.\\
    3.4.2 A dog and cat are \emph{implying} together on an orange couch.\\
    3.4.3 A dog and cat are \emph{cleaning} together on an orange couch.\\[1ex]
    3.5.1 A \emph{cat} and a dog rest together on a hideous orange couch.\\
    3.5.2 A \emph{settlement} and a dog rest together on a hideous orange couch.\\
    3.5.3 A \emph{cuddle}and a dog rest together on a hideous orange couch.\\
    \end{tabular}}
    \end{minipage}
    \vspace{4ex}

    \begin{minipage}[t]{.4\textwidth}\vspace{0pt}%
    \includegraphics[width=\textwidth]{Images/000000453926.jpg} 
    \end{minipage} 
    \begin{minipage}[t]{.58\textwidth}\vspace{0pt}%
    {\small
    \begin{tabular}{l}
    4.1.1 A \emph{plate} that has a dessert on it\\
    4.1.2 A \emph{lobster} that has a dessert on it.\\
    4.1.3 A \emph{glaze} that has a dessert on it.\\[1ex]
    4.2.1 A chocolate and fudge dessert on layered pastry is on a red \emph{plate}.\\
    4.2.2 A chocolate and fudge dessert on layered pastry is on a red \emph{mush}.\\
    4.2.3 A chocolate and fudge dessert on layered pastry is on a red \emph{slop}.\\[1ex]
    4.3.1 A \emph{stack} of pancakes on a plate with creme and chocolate.\\
    4.3.2 A \emph{restaurant} of pancakes on a plate with creme and chocolate.\\
    4.3.3 A \emph{pot} of pancakes on a plate with creme and chocolate.\\[1ex]
    4.4.1 a red \emph{plate} with some chocolate and whip cream desert.\\
    4.4.2 a red \emph{barrier} with some chocolate and whip cream desert.\\
    4.4.3 a red \emph{prod} with some chocolate and whip cream desert.\\[1ex]
    4.5.1 A small cake is \emph{covered} in frosting on a plate.\\
    4.5.2 A small cake is \emph{triggered} in frosting on a plate.\\
    4.5.3 A small cake is \emph{competed} in frosting on a plate.\\
    \end{tabular}}
    \end{minipage}
    \vspace{4ex}

    \begin{minipage}[t]{.4\textwidth}\vspace{0pt}%
    \includegraphics[width=\textwidth]{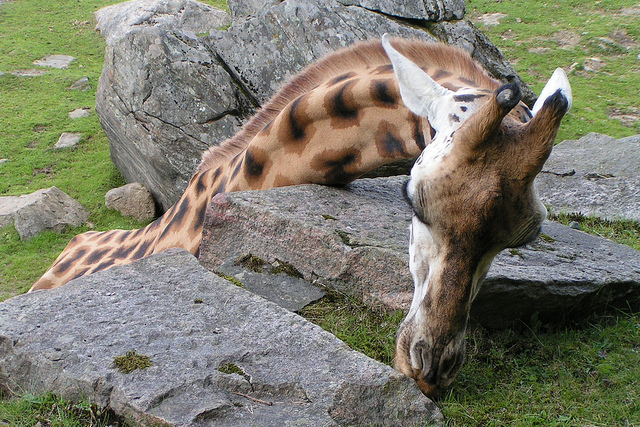} 
    \end{minipage} 
    \begin{minipage}[t]{.58\textwidth}\vspace{0pt}%
    {\small
    \begin{tabular}{l}
    5.1.1 A \emph{giraffe} looking for food between large rocks.\\
    5.1.2 A \emph{badger} looking for food between large rocks.\\
    5.1.3 A \emph{dock} looking for food between large rocks.\\[1ex]
    5.2.1 A giraffe \emph{rests} it's neck on a bunch of rocks.\\
    5.2.2 A giraffe \emph{thrust} it's neck on a bunch of rocks.\\
    5.2.3 A giraffe \emph{flew} it's neck on a bunch of rocks.\\[1ex]
    5.3.1 \emph{A giraffe} laying down with his head on the rocks.\\
    5.3.2 \emph{An eagle} laying down with his head on the rocks.\\
    5.3.3 \emph{A barb} laying down with his head on the rocks.\\[1ex]
    5.4.1 A \emph{giraffe} reaching between two rocks to obtain grass.\\
    5.4.2 A \emph{lizard} reaching between two rocks to obtain grass.\\
    5.4.3 A \emph{coat} reaching between two rocks to obtain grass.\\[1ex]
    5.5.1 A \emph{giraffe} reaching its head above some rocks to a grass area.\\
    5.5.2 A \emph{turtle} reaching its head above some rocks to a grass area.\\
    5.5.3 A \emph{bat} reaching its head above some rocks to a grass area.\\
    \end{tabular}}
    \end{minipage}
    \vspace{4ex}

\end{document}